\documentclass[10pt,letterpaper,twocolumn]{article}

\usepackage{amsmath} 
\usepackage{multirow,tabularx}
\usepackage{graphicx}
\usepackage{float}
\usepackage[utf8]{inputenc}
\usepackage{eurosym}
\usepackage{accents}
\usepackage{glossaries}
\newacronym{NASA}{NASA}{National Aeronautics and Space Administration}
\newacronym{FSM}{FSM}{Finite State Machine}

\usepackage{times}
\usepackage{titlesec}

\titleformat{\section}{\normalfont\normalsize\bfseries}{\thesection}{1em}{}
\titlespacing*{\section}{0pt}{1ex}{1ex}
\titleformat{\subsection}{\normalfont\normalsize\bfseries}{\thesubsection}{1em}{}
\titlespacing*{\subsection}{0pt}{1ex}{1ex}
\titleformat{\subsubsection}{\normalfont\normalsize\bfseries}{\thesubsubsection}{1em}{}
\titlespacing*{\subsubsection}{0pt}{1ex}{1ex}

\title{\large \textbf{HUMAN-ROBOT COLLABORATION IN MICROGRAVITY: THE OBJECT
  HANDOVER PROBLEM} \\ \vspace{1ex}
  \normalsize Virtual Conference 19–23 October 2020 \vspace{-2.5ex}}

\author{\normalsize Adriana Fernandes$^{1}$, Rodrigo Ventura$^{1}$ \vspace{-2ex}}

\date{\normalsize $^{1}$Institute for Systems and Robotics, Instituto Superior Técnico, Av. Rovisco Pais 1, 1049-001 Lisboa,
  Portugal, E-mail: adrianafernandes@tecnico.ulisboa.pt \\
  $^{2}$Institute for Systems and Robotics, Instituto Superior Técnico, Av. Rovisco Pais 1, 1049-001 Lisboa,
  Portugal, E-mail: rodrigo.ventura@isr.tecnico.ulisboa.pt \vspace{-3ex}}

\begin{document}

\maketitle
\thispagestyle{empty}
\pagestyle{empty}

\begin{abstract}
  Collaborative space robots are an emerging technology with high
  impact as robots facilitate servicing functions in collaboration
  with astronauts with higher precision during lengthy tasks, under
  tight operational schedules, with less risk and costs, making them
  more efficient and economically more viable. However, human-robot
  collaboration in space is still a challenge concerning key issues in
  human-robot interaction, including mobility and collaborative
  manipulation of objects on a microgravity environment.  In this
  paper we formulate an algorithm that enables a free-flyer robot,
  equipped with a manipulator, to perform an object handover between a
  human and a free-flyer robot, in a microgravity environment. To
  validate and evaluate this algorithm, we present a systematic user
  study with the goal of understanding the subjective outcome effects
  of a rigid and compliant impedance robot behavior during the
  interaction.  The results showed that the rigid behavior was overall
  more preferable and registered higher transfer success during the
  tasks.
\end{abstract}

\section{INTRODUCTION}

Collaborative space robotics comprises the development of autonomous
robots that are able to operate in microgravity environments
facilitating manipulation, assembling or servicing functions in
collaboration with astronauts. One of the reasons why space robots
became progressively more relevant is their potential to offload tasks
from astronauts, and thus reduce the crew size in-orbit. Robots also
operate with extreme high precision and repeatability, being crucial
on a demanding and safety-critical environment such as space.
Additionally, the space environment is inherently risky, due to, e.g.,
radiation and space debris, posing an ever-present danger to
astronauts.

Many advances were made regarding collaborative space robotics
challenges but, to the best of the authors knowledge, no prior work on
the field of object handover between humans and robots in microgravity
was found in the literature. This problem differs from the terrestrial
one by the absence of any cue to the human caused by weight of the
object being handed over.  Nevertheless, it is important to state
relevant research in terrestrial object handovers of the following
three phases: approach, transfer and retraction. Regarding the
approach phase, Cakmak in \cite{handover17} showed that robot's
postures with an extended arm were most frequently classified handing
over and Koay in \cite{handover19} concluded that the robot should
approach the user from the front. Furthermore, Aleotti in
\cite{handover9} stated that robots should take into consideration how
the human will grasp the object and thus robots should approach the
user with the easiest part to grasp of the object. Regarding the
transfer phase, Edsinger in \cite{handover8} found that humans will
pose an object in the robot’s stationary hand regardless of the
robot’s hand pose. Regarding the communication intent, Strabala in
\cite{handover11} claims that special signals can be used when the
human and robot share the meaning of these signals in a common
ground. Concerning the decision of releasing or grasping an object,
Edsinger in \cite{handover8} monitored the velocity of the robot's
end-effector.  To achieve a dynamic handover, Kupcsik’s studied a
Cartesian impedance control approach \cite{handover4} and Kumagai in
\cite{handover22} presented an implementation of a human-inspired
handover controller on a robot with compliant under-actuated
hands. Concerning the retraction phase, Strabala in \cite{handover11}
stated that after transferring the entire object load, often the
receiver will retract, indirectly signalling the giver that the
handover is complete.

The goal of the this paper is to formulate an algorithm that enables a
free-flyer robot equipped with a manipulator to perform an object
handover with a human on a successful, fluent, and dynamic manner in a
microgravity environment.
The proposed algorithm is based on a Finite State Machine (FSM)
coordinating the behavior of the robot during the following handover
phases: approach, transfer, and retraction. A standard impedance controller
was designed and implemented for the transfer phase, in which the
object is grasped simultaneously by the human and robot.

For the validation and evaluation of the algorithm, a systematic user
study was conducted, using an user interaction interface that was
developed, based on a virtual reality environment. This environment
uses the open-source Astrobee simulation
platform~\cite{astrobeesoft}. The virtual reality environment uses a
Leap Motion device for tracking in real time a human hand. This
environment was used on a systematic user study.

The present paper is structured as follows: The Handover Algorithm
Formulation is described in Section \ref{formulation}, including the
derivation of the impedance controller. Furthermore, the
Implementation and Results are presented in Section
\ref{sec:impresults}, the User Study is shown in Section
\ref{userstudysec} and Section \ref{conclusionsection} includes
concluding remarks with future work references.

\section{HANDOVER ALGORITHM FORMULATION}
\label{formulation}

The proposed algorithm comprises two levels of abstraction: a
\acrfull{FSM} modeling the multiple handover phases, and a set of
motion controllers, one for each phase.

\subsection{\textbf{Robot-to-Human Handover}}

A robot-to-human object handover task aims to achieve an object transfer from a robot to a human where the robot acts as the giver and the human as the receiver. In this manner, a sequence of states and transitions of the FSM proposed were selected as Fig. \ref{fig:H-Rscheme} displays in green. Initially and assuming that the intention to perform a handover has already been established, the first state involves the opening of the gripper, followed by the movement to the object location which is assumed to be known by the robot. Upon arrival, the object must be grasped. It is important to refer that no specific grasping algorithm was designed given that the object grasping field was considered a sub-problem out of the scope of this paper. The next states involves the robot's movement into the handover pre-assigned location. Moreover, the robot should approach the user from the front as this angle provides him/her the most visibility of the robot's motion  \cite{handover19}. Furthermore, the robot's arm should be extended \cite{handover17}. With the aim of delivering the object in a dynamic and fluent manner, an impedance control-based approach must be activated. This approach implements a dynamic response between the environment acting on the robot's manipulator structure and its motion. This module is further analysed in section \ref{impedancecontrolform}. The impedance control (IC) activation is followed by the state regarding the user signalling in which the robot should communicate to the user that it is ready to deliver the object \cite{handover11}. Another relevant stage of the transfer phase is the robot's decision concerning the appropriate time of releasing the object. Following the work developed by Edsinger in \cite{handover8}, the robot's end-effector velocity was monitored. In this manner, if the end-effector velocity is higher then the defined threshold, $\alpha$, and the robot is grasping the object, the user's receiving intention is detected and the robot will open the gripper. The retraction phase is the last phase of the handover sequence. If the object has been delivered, the robot must switch off the formulated impedance control and move away from the handover location.

\subsection{\textbf{Human-to-Robot Handover}}
\begin{figure}[h]
  \centering
  \includegraphics[width=0.49\textwidth]{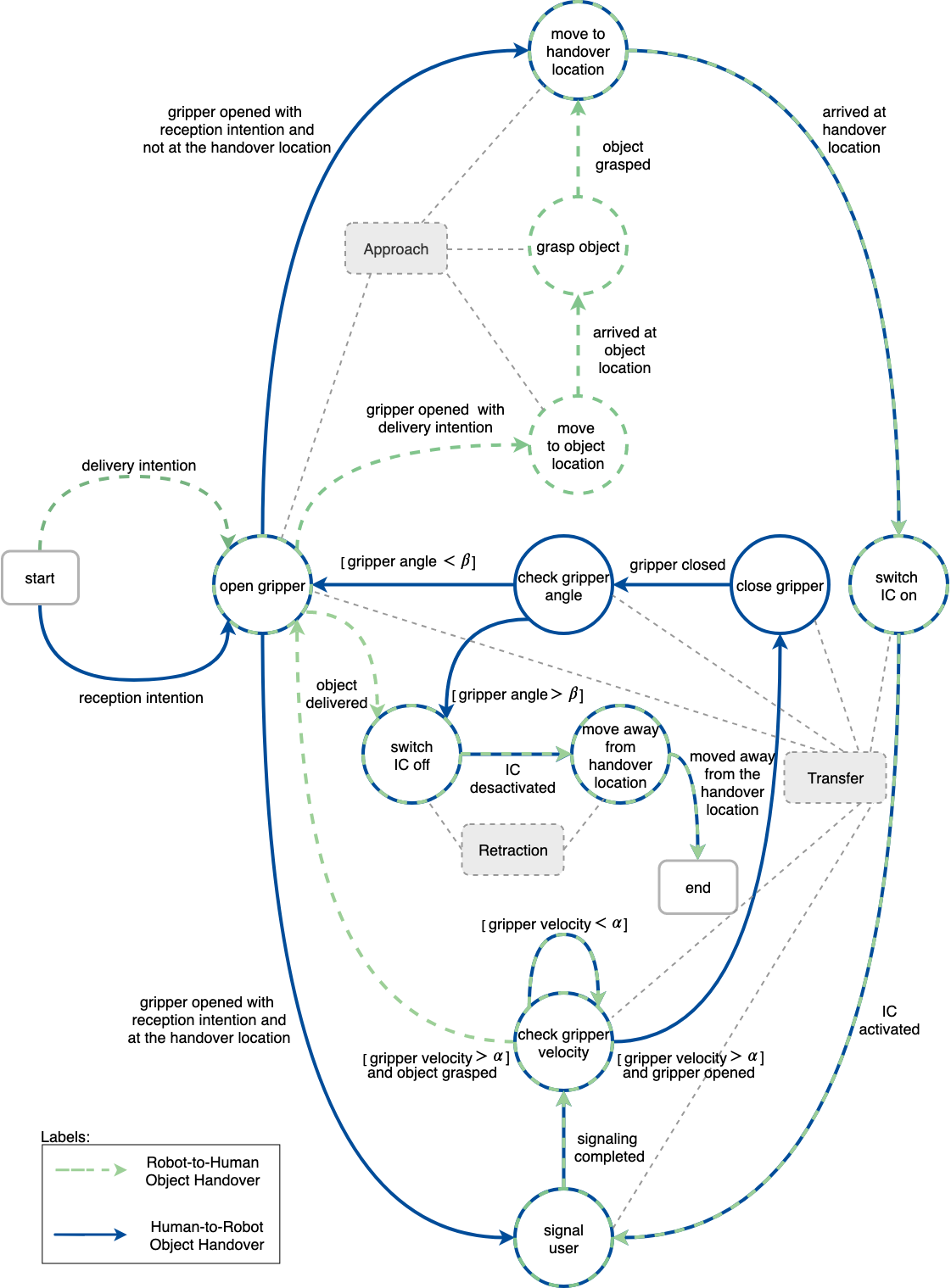}
  \caption{State-machine based algorithm sequence, in blue, regarding a human-to-robot handover. The three phases on the handover are also presented.}
  \label{fig:H-Rscheme}
\end{figure}

 In a human-to-robot object handover task, the robot is the receiver and the human performs the giver role. A sequence of transitions and states of the proposed FSM describes a human-to-robot handover and it is presented in Fig. \ref{fig:H-Rscheme} in blue. The approach phase is initiated without the object and it is assumed that the robot already acknowledges the intention of receiving an object. As in the previous task, the robot's arm should be extended in the approach phase \cite{handover17}. Furthermore, the opened gripper during the approach stage emphasis that intention. Upon arrival to the handover location, the same IC approach used on the robot-to-human handover task must be activated and the robot must signal the user. Furthermore, the robot must detect that the object has been placed in its end-effector. The work developed by Edsinger in \cite{handover8} was again taken into consideration. Upon closing its gripper, the robot must verify the object reception success. This can be done by checking the resulting grasp aperture: if it is positive and above a threshold, $\beta$, then the gripper is assumed to be wrapped around an object and the retraction phase initiates, otherwise the robot must re-open the gripper and signal the user, showing the acknowledgement of the a failed transfer. Lastly, in the retraction phase, the IC must be switched off and the robot must move away from the handover location.

\subsection{\textbf{Impedance Control}}
\label{impedancecontrolform}
A fluent and dynamic human-robot handover may be achieved due to the robot's adaptability to the task conditions, environmental constraints and perturbations instead of simply controlling its position, in which the robot is seen as an isolated system. As a result, impedance control was selected as the controlling approach for the transfer phase of the proposed FSM-based handover algorithm. This section aims to formulate an impedance controller that generates a dynamical relationship between a free-flyer robot manipulator and external forces acting on it. This formulation is adapted to this paper goals from the research developed by Lippiello and Ruggiero in \cite{lippielloruggiero}.

\subsubsection{Kinematic Model}

 The manipulator consists of $n$ rigid links connected by joints $q_l$, with $l = 1,2,3, ..., n$. Moreover, the inertial frame is denoted by ${\Upsilon_i}$, the body-fixed reference frame placed at the spacecraft center of mass by ${\Upsilon_b}$ and the end-effector coordinates attached to the interaction point of the manipulator by ${\Upsilon_e}$.  Furthermore, the absolute position of ${\Upsilon_b}$ with respect to ${\Upsilon_i}$ is described as $p^i_b = [x_b\;y_b\;z_b]^T$ and the system attitude is expressed in roll-pitch-yaw Euler angles being denoted by $\phi^i_b = [\varphi_b\; \theta_b \; \psi_b]^T$. Additionally, the absolute transitional velocity of ${\Upsilon_b}$ is represented by $\Dot{p}^i_b$ and $\Dot{p}^b_b$, with respect to ${\Upsilon_i}$ and to ${\Upsilon_b}$, respectively. Regarding the absolute rotational velocity, $\omega^i_b$ refers to the absolute rotational velocity of the vehicle and $\omega^b_b$ denotes the absolute rotational velocity with respect to ${\Upsilon_b}$.
If the rotation matrix of frame ${\Upsilon_b}$ with respect to frame ${\Upsilon_i}$ is defined by $R^i_b$, the spacecraft linear velocity representation in ${\Upsilon_b}$ coordinates is transformed to its representation in ${\Upsilon_i}$ coordinates from:
\begin{equation}
\centering
    \Dot{p}^i_b = R^i_b \; \Dot{p}^b_b
    \label{pRpdot}
\end{equation}
Moreover, if the transformation matrix between the time
derivative of $\phi^i_b$ and $\omega^i_b$ is defined by $N^i_b$, the transformation of the free-flyer absolute rotational velocity is obtained as:
\begin{equation}
\centering
    \omega^i_b = N^i_b \; \Dot{\phi}^i_b
    \label{wSphidot}
\end{equation}
From (\ref{pRpdot}) and (\ref{wSphidot}) holds:
\begin{equation}
\centering
    \omega^b_b = (R^i_b)^T \; \omega^i_b = (R^i_b)^T \; N^i_b \; \Dot{\phi}^i_b = Q^i_b \; \Dot{\phi}^i_b
    \label{wSphidot2}
\end{equation}
with $Q^i_b=(R^i_b)^T \; N^i_b$ being the mapping of the time derivative of $\phi^i_b$ into the body absolute rotational velocity with respect to ${\Upsilon_b}$. The transformation equations (\ref{pRpdot})-(\ref{wSphidot2}) are valid as long as the matrices $R^i_b$, $N^i_b$, and $Q^i_b$ are non-singular.  Furthermore, direct kinematics of the spacecraft are defined by the following transformation matrix:
\begin{equation}
\centering
    K^i_b(p^i_b, {\phi}^i_b) = \begin{bmatrix}
  R^i_b({\phi}^i_b) &
   p^i_b \\
   0_{1\times3} &
   1 
   \end{bmatrix}
    \label{Kuav}
\end{equation}
where, $0_{1\times3}$ is a $(1\times3)$ vector composed only by zeros. Furthermore, direct kinematics of the manipulator with respect to ${\Upsilon_b}$ are expressed as:
\begin{equation}
\centering
    K^b_e(q) = \begin{bmatrix}
  R^b_e(q) &
   p^b_e(q) \\
   0_{1\times3} &
   1 
   \end{bmatrix}
    \label{Kmanipulator}
\end{equation}
with $q$ describing the $(n\times1)$ vector of the robot manipulator
joints variables, $R^b_e$ the rotation matrix between ${\Upsilon_e}$ and ${\Upsilon_b}$ and $p^b_e = [x^b_e\;y^b_e\;z^b_e]^T$ the position of the end-effector with respect to ${\Upsilon_b}$. Combining (\ref{Kuav}) and (\ref{Kmanipulator}):
\begin{equation}
\centering
    K^i_b \; K^b_e = K^i_e(\xi) 
    \label{K_E}
\end{equation}
where $\xi= [{p^i_b}^T\; {\phi^i_b}^T \; q_1 \; ... \; q_n]^T$  is the $((6 + n) \times 1)$ generalized vector of the system joints variables.
Moreover, the end-effector absolute position with respect to the inertial frame is defined as $x = [{p^i_e}\; {\phi^i_e}]$ where $p^i_e = [x_e\;y_e\;z_e]^T$ and the manipulator's attitude is also expressed in roll-pitch-yaw Euler angles being denoted by $\phi^i_e = [\varphi_e\; \theta_e \; \psi_e]^T$ with respect to ${\Upsilon_i}$. The vector of absolute generalized velocity of the manipulator’s end-effector can consequently be expressed as $\Dot{x} = [{\Dot{p}^i_e}\; {\dot{\phi}^i_e}]$. 
The transformation between $\Dot{x}$ and the time derivative of the system generalized joints variables can be written as:
\begin{equation}
\centering
    \Dot{x} = J \; \Dot{\xi}
    \label{x=Jqsi}
\end{equation}
where $J$ is the so called Jacobian $(6\times (6 + n))$ matrix of the system. 
Moreover, time deriving (\ref{x=Jqsi}) yields:
\begin{equation}
\centering
    \ddot{x} = J \; \ddot{\xi} + \Dot{J} \; \Dot{\xi}
    \label{xdotdotJacobian}
\end{equation}

\subsubsection{Dynamic Model}
According to the \textit{Euler-Lagrange} formulation, the Lagrangian of a mechanical system in the absense of gravity (and of any other source of potential energy) is simply given by $L=T$, where $T$ is the total kinetic energy. The solution is given by the well-known Euler-Lagrange equation:
\begin{equation}
\centering
   \frac{d}{dt} \frac{\partial L}{\partial \Dot{\xi_i}} - \frac{\partial L}{\partial {\xi_i}} = u_i
    \label{lagrange}
\end{equation}
with $i$ describing the $i$-th generalized coordinate of $\xi$ and assuming values of $i = 1, ..., ((6+n))$. The $i$-th generalized force is represented as $u_i$. The total kinetic energy of the system being studied is composed by the energy contributions concerning the motion of the spacecraft, $T_b$ and the energy associated with motion of each link of the manipulator, $T_{l_i}$, as express in (\ref{totalkineticsystem}).
\begin{equation}
\centering
   T = T_b + \sum_{i=1}^{n} T_{l_i} = \frac{1}{2}\;{\Dot{\xi}}^T \; B \; \Dot{\xi}
    \label{totalkineticsystem}
\end{equation}

with $B$ being an $((6+n)\times(6+n))$ symmetric and positive inertia matrix. Lastly, computing the Lagrange equation, the dynamics of the system in the generalized joint space are given by:
\begin{equation}
\centering
   B(\xi) \; \ddot{\xi} + C(\xi, \Dot{\xi}) \; \Dot{\xi} =  u + u_{ext}
    \label{systemlagrangefinal}
\end{equation}
where $u$ describes the generalized input forces vector $((6+n)\times1)$ and $u_{ext}$ represents the external generalized forces vector at a joint level,  $((6+n)\times1)$. Furthermore, $C$ is an $((6+n)\times(6+n))$ matrix that encompasses the Coriolis and centrifugal terms.

\subsubsection{Control Law}

Let $\Ddot{x}_d$,  $\dot{x}_d$ and ${x_d}$ be the end-effector desired \textit{rest} acceleration, velocity and position, respectively, and the actual position error as $\Tilde{x} = x_d - x$.  Moreover, during the transfer phase on the the handover tasks formulated it is assumed that $\ddot{x}_d = 0$ and $\dot{x}_d = 0$.  Given these considerations, a suitable law control can be designed:
\begin{equation}
\centering
 u = J^T \;(- K_B \; \dot{x} + K_D \; \Tilde{x} )
    \label{ugeral}
\end{equation}
With $K_D$ and $K_B$ representing the $((6+n)\times(6+n))$ symmetric and positive definite matrices of the chosen stiffness and damping, respectively. It is important to refer that these matrices can be tuned to the desired system's behavior. Finally, substituting (\ref{ugeral}) into (\ref{systemlagrangefinal}) and considering (\ref{x=Jqsi}) and (\ref{xdotdotJacobian}), the joint space dynamics can be expressed in terms of the manipulator’s end-effector configuration, $x$, in the inertial Cartesian coordinates representing an impedance dynamic model as presented in (\ref{dynamicrelat}).
\begin{equation}
\centering
 B_x \; \ddot{x} + (C_x + K_B) \; \dot{x} - K_D \; \Tilde{x} =  f_{ext}
\label{dynamicrelat}
\end{equation}
with $f_{ext}$ representing the vector $((6+n)\times1)$ of the external generalized forces at the Cartesian coordinate level and $B_x$ and $C_x$ describing the inertia and Coriolis matrices with respect to the $x$ variable:

\begin{subequations}
\begin{equation}
\centering
B_x = J(\xi)^{-T} \; B(\xi) \; J(\xi)^{-1}
\end{equation}
\begin{equation}
C_x = J(\xi)^{-T} \; ( C(\xi,\dot{\xi}) - B(\xi) \; J(\xi)^{-1} \;\dot{J}(\xi) ) \; J(\xi)^{-1}
\end{equation}
\end{subequations}

with $^{-T}$ being the inverse of transpose.

Summarizing, a control law was designed for the transfer phase of the formulated FSM-based handover algorithm given the kinematics and dynamics of a microgravity free-flying robot equipped with a manipulator.

\section{IMPLEMENTATION AND RESULTS}
\label{sec:impresults}

Given the availability of an open-source Astrobee software platform designed to conduct research \cite{astrobeesoft}, this free-flyer robot simulator was used as an implementation platform to showcase and verify the formulated handover algorithm. In the future, this formulation and implementation can validated on the Astrobee aboard the ISS.
The implement was done using Python 3.0, Ubuntu 16.04 LTS and ROS Kinetic.

\subsection{Impedance Control Validation}

\begin{figure*}[h]
 \centering
  \includegraphics[width=\linewidth]{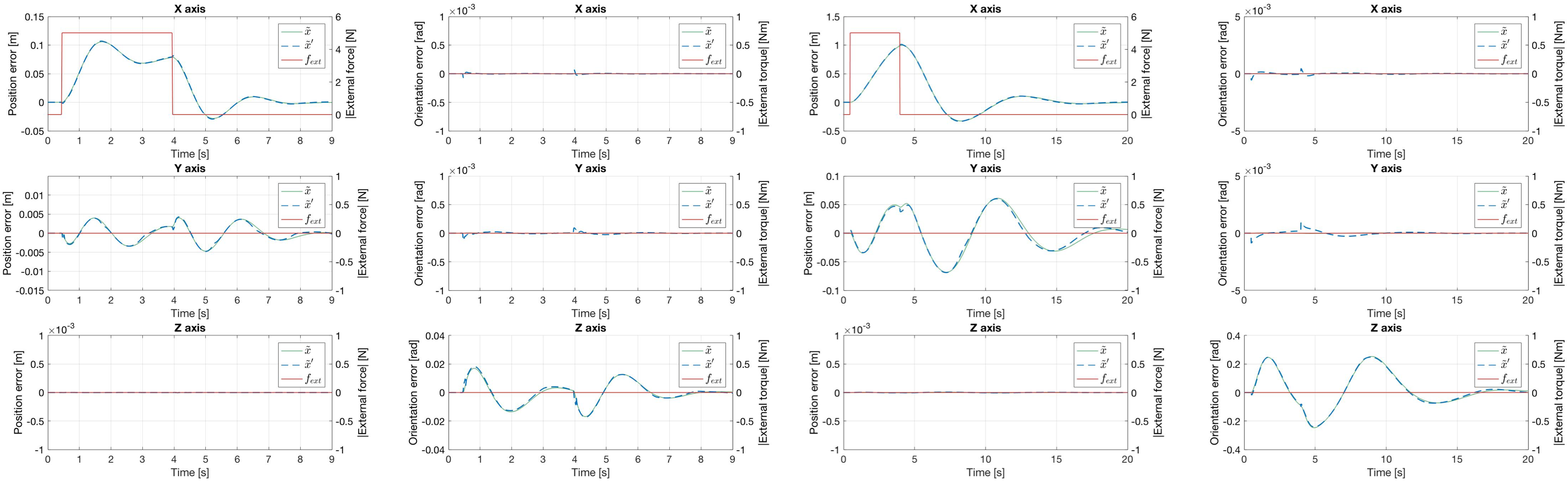}
  \caption{Actual and computed (from impedance model) end-effector position and orientation error, with a rigid behavior (6 figures on the left) and compliant behavior (6 figures on the right). A $5N$ force was applied on the X axis.}
  \label{impedancevali}
\end{figure*}

Regarding the impedance control module, it is important to refer that
although an impedance control formulation for a free-flyer robot
equipped with a manipulator was described on section
\ref{impedancecontrolform}, the Astrobee's Gazebo
simulator\footnote{https://www.nasa.gov/astrobee} presented
considerably small tolerances for the joints state goals. Thus, it
does not allow the movement of the arm links for small controlled
angles and the joint variables are assumed to be fixed for this
implementation. In this manner:
\begin{equation}
\Dot{\xi}= [{\Dot{p}^i_b}^T\; {\Dot{\phi}^i_b}^T \; 0 \; ... \; 0]^T
\end{equation}
Moreover, two types of impedance behaviors were implemented: rigid and
compliant. These behaviors can be defined by tuning the values of the
matrices $K_D$ and $K_B$ in (\ref{ugeral}). In order to obtain the
formulated impedance controller validation during interaction, several
generalized external forces, $f_{ext}$, were applied to the robot's
end-effector, for both behavior study cases. Furthermore, with the aim
of validating the dynamic impedance model proposed, the expected
values of the end-effector position and orientation error,
$\Tilde{x}'$, were computed from (\ref{dynamicrelat}) given the
end-effector simulated acceleration, $\ddot{x}$, the end-effector
simulated velocity, $\Dot{x}$, the $B_x$, $C_x$, $K_D$ and $K_B$
matrices and the $f_{ext}$, as follows:
\begin{equation}
\centering
 \Tilde{x}' = K_D^{-1}\;\left[B_x \; \ddot{x} + (C_x + K_B) \; \dot{x}  \; -  f_{ext}\right]
\label{xcalculted}
\end{equation}
The  end-effector position and orientation errors, $\Tilde{x}'$, as well as the actual simulated end-effector position and orientation error, $\Tilde{x}$, are presented each axis in Fig.~\ref{impedancevali} for the rigid behavior and for the compliant behavior case. Additionally, the external generalized forces generated for each simulation are represented.

Analysing the presented figures, several conclusions can be drawn. The first one concerns the clear motion distinction between the two behaviors: a higher stiffness value in (\ref{dynamicrelat}) generates a rigid behavior where the end-effector tends to reach the desired state with a lower position/orientation error and a lower stiffness value generates robot's motion passively to the external perturbation, diverging more from the desired state. Furthermore, the robot not only reaches for desired/rest state for both behaviors while an external force acts on the end-effector, but also in the absence of external perturbations. Lastly, the end-effector behaves accordingly to desired impedance model expressed by (\ref{dynamicrelat}). This can be concluded due to the overlap of the actual simulated end-effector error motion, $\Tilde{x}$, and the computer end-effector error motion from the impedance model, $\Tilde{x}'$.

\subsection{Human Interaction Implementation}

\begin{figure*}[h]
\centering
  \includegraphics[width=\linewidth]{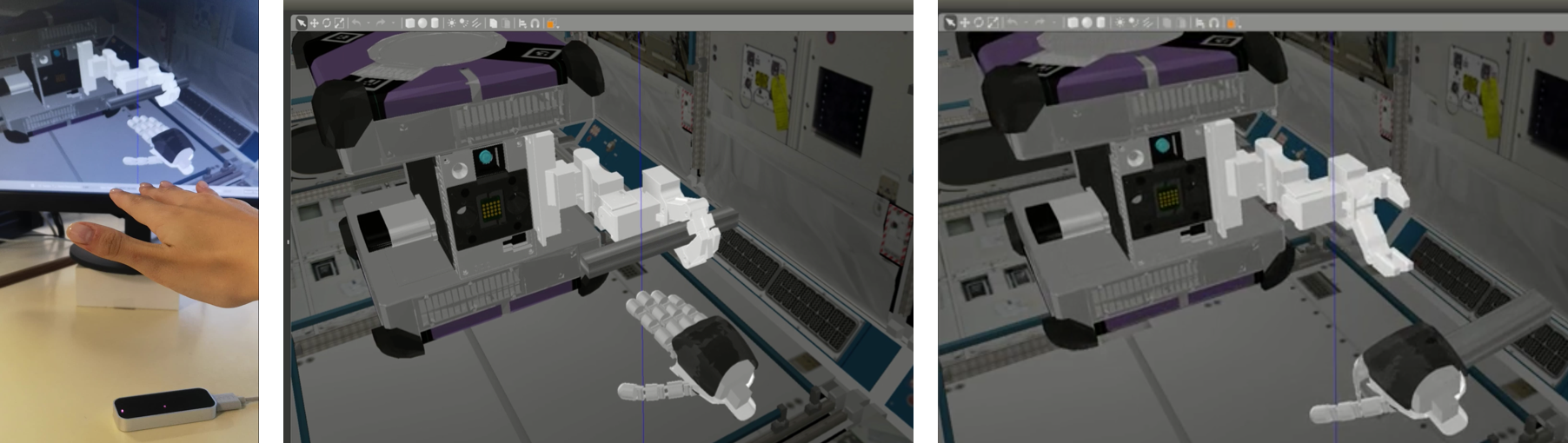}
  \caption{Virtual reality simulation environment. Left: human hand
    being tracked by the Leap Motion device (bottom), and the
    simulated hand on the computer screen. Center and right: Astrobee
    simulation environment showing the robot handing over a
    cylindrical bar to the simulated hand.}
  \label{Astrobeeandhand}
\end{figure*}

The algorithm proposed assumes that two agents are involved in the
handover: a robot and a human. Thus, a simulated hand model controlled
via a real user hand was implemented. The main requirement concerning
the simulated hand model is its the ability to mimic a human hand in
terms of its degrees of freedom. The iCub robot hand ended up being
selected and integrated into the Gazebo simulation, as no other more
photo realistic simulated human hand was found by the authors. To
control the simulated hand, the values of the real user hand
position/orientation and fingers were tracked via a Leap Motion device
and integrated on the interface. Fig. \ref{Astrobeeandhand} displays
the user hand, the Leap Motion device and the simulation environment.

It should be acknowledge that the absense of any haptic feedback on
the user hand is a limitation of this study. Thus, the visual feedback
of the virtual environment is the only modality the user has to
perceive the environment state.

\subsection{{Handover Algorithm Validation}}

The algorithm was validated in both proposed tasks. In
the case of a robot-to-human handover and as formulated, the robot
initiated the handover with a closed gripper and without the
object. It then opened the gripper, got closer to the object and
grabbed it. Following, the robot moved to the handover location. After
activating the IC, the robot was ready to deliver the object and the
Transfer phase initiated in which the Astrobee signaled the user using
its flashlight and waited for the gripper velocity threshold. When
this occurs, the robot opened its gripper and the object is
transferred to the user. Lastly, both moved away from the handover
location.  In the case of a human-to-robot handover, the robot
initiated the handover with a closed gripper and without the object as
the user is grabbing it. It then opened the gripper and moved to the
same handover location. After activating the IC, the robot was ready
to receive the object and thus, the Transfer phase initiated in which
the robot signaled the user and waited to detect the object placement
on its end-effector. When this occurs, the transfer of the object was
performed and the user hand model and the Astrobee moved away from the
handover position. Furthermore, the failure detection module was
successfully validated given that the robot acknowledge a test failed
transfer by re-opening the gripper and re-signaling the user.  Lastly,
during the transfer for both tasks the gripper motion was in
accordance to the impedance control results, given that, for a similar
user external interaction, the gripper’s movement was minimum for a
rigid behavior and it moved passively to simulated user hand for the
compliant case.

\section{USER STUDY}
\label{userstudysec}

\begin{figure*}[h]
  \includegraphics[width=\linewidth]{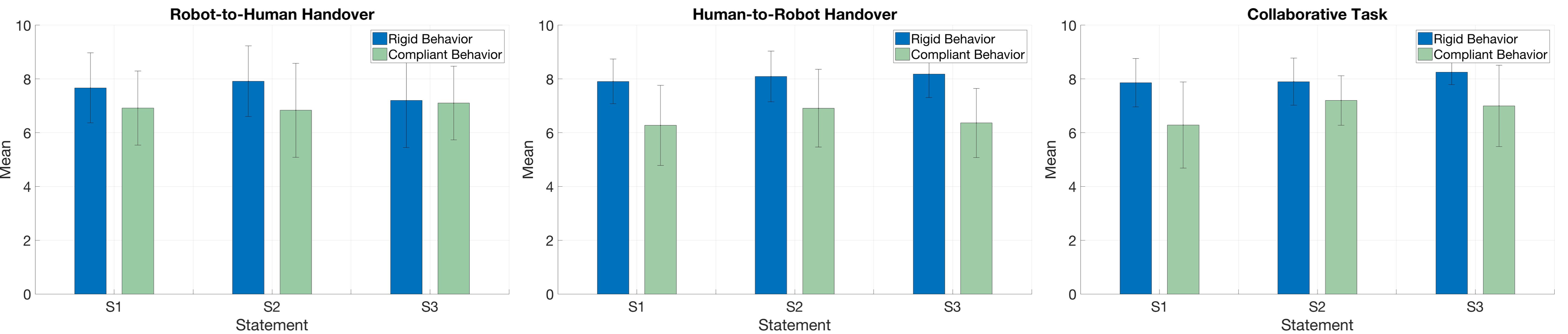}
  \caption{Representation of the mean and standard deviation of the questionnaire's results for the performed tasks.}
  \label{userstudyres}
\end{figure*}

Based on the results of Kupcsik’s study \cite{handover4} it is known
that for static handover tasks using cartesian compliant control,
compliance parameters are less important for success and high
stiffness is always preferred and highly rated. Gasparri in
\cite{userstudy2} shows that when using impedance handover dynamics
the optimal manipulator stiffness is high in the case of perfect
knowledge of the framework. In this sense, the systematic user study
aimed to explore the subjective outcomes effects on the user
concerning the implemented robot behaviors. Furthermore, the handover
success of the two behaviors was also studied. In this sense, two
hypothesis were in advance proposed for the experimental study:
\begin{description}
\item[\textbf{H1}:]  \emph{The impedance control parameters will affect the
  participant’s perception of the object handover task with high
  stiffness (rigid behavior) being the most fluent, desirable and
  cooperative and low stiffness (compliant behavior) the less fluent,
  desirable and cooperative;}
\item[\textbf{H2}:] \emph{the impedance control parameters will affect the object handover task success with high stiffness (rigid behavior) being the most successful and low stiffness (compliant behavior) the less successful.}
\end{description}

Ten people with ages between 21 and 30 participated in this experiment (6 female and 4 male).
Initially each participant performed different manoeuvres of their
choice with the simulated hand for 10 minutes. The second section of
the experiment was the handover tasks: robot-human, human-robot
handover and a collaborative task that encompassed both
handovers. Moreover, the controller parameters conditions were
adjusted in order to achieve rigid behavior or compliant behavior. The
study involved 12 rounds of interaction for each participant – two for
each experimental condition with randomized controlled trials. After
each round of interaction, participants filled out a questionnaire
giving a score between 1 (fully disagree) and 9 (fully agree)
\cite{userstudy5} to three statements regarding their perception of
the handover. In particular three scales were used — fluency
\cite{userstudy3} and \cite{userstudy7}, satisfaction
\cite{userstudy5} and team work \cite{userstudy3}. The statements were
the following:
\begin{description}
\item[\textbf{S1}:] ``The robot contributed for the fluency of the interaction.'';
\item[\textbf{S2}:] ``I was satisfied with the interaction.'';
\item[\textbf{S3}:] ``The robot was committed to the task.''
\end{description}
Additionally, the number of non-successful object handover in the three tasks were registered.
\subsection{Results}
The questionnaire's results are presented in Fig. \ref{userstudyres}, for the proposed tasks. Additionally, Table \ref{fig:sucess} displays the total number of failed transfers for each task.

\begin{table}[h]
\centering
 \caption{Total number of failed handovers from a total of 12 runs, on the three performed tasks.}
 \footnotesize
 \renewcommand{\arraystretch}{1.5}
 \vspace{3ex}
\begin{tabular}{cc|c|c|}
  \cline{3-4}
 &  & \begin{tabular}[c]{@{}c@{}}Rigid \\ Behavior\end{tabular} & \begin{tabular}[c]{@{}c@{}}Compliant\\ Behavior\end{tabular} \\ \hline
\multicolumn{1}{|c|}{\multirow{3}{*}{Task}} & Robot-to-Human Handover & 3 & 3 \\ \cline{2-4} 
\multicolumn{1}{|c|}{} & Human-to-Robot Handover & 0 & 2 \\ \cline{2-4} 
  \multicolumn{1}{|c|}{} & Collaborative & 2 & 3 \\
  \hline
\end{tabular}%
  \label{fig:sucess}
\end{table}

Concerning the user's responses to the proposed statements, the
results indicate that in a robot-to-human handover scenario users
tended to perceive higher fluency for a rigid behavior (p-value
$=0.1195$) and tended to be more satisfied with the interaction also
for the rigid behavior (p-value $ =0.0589$). Moreover, no substantial
difference between both behaviors was felt regarding the robot
commitment to the task (p-value $=0.8880$) and thus concerning the
cooperation perceived. Additionally, results shows higher distinction
between the answers regarding the two behaviors in the human-to-robot
handover scenario, as users perceived more fluency, satisfaction and
cooperation for a rigid behavior, with statistical significance
(p-value of $0.0132$, $0.0401$ and $0.0057$, respectively). As
expected, for the collaborative task, the results were also higher for
the rigid behavior concerning S1, S2 and S3, with statistical
significance (p-value of $0.0375$, $0.0445$, $0.0492$,
respectively). Lastly, more total successful handovers were performed
for the rigid behavior. Summarizing, the results indicate that H1 was
verified for the first two factors and the handover success data
supported H2 for the human-to-robot object handover and collaborative
task. In this sense, the users perceived an overall more fluent,
dynamic and successful object handover for the rigid behavior.

\section{CONCLUSIONS AND FUTURE WORK}
\label{conclusionsection}

This paper formulated and validated an algorithm enabling a free-flyer
robot to perform an object handover with a human in a microgravity
environment, in a dynamic, fluent and successful manner, combining a
FSM and an impedance controller. Furthermore, a virtual reality user
interaction interface on a realistic robotic simulator was developed
and a systematic user study was conducted where results showed that
the rigid behavior was overall more preferable and registered higher
transfer success during the proposed tasks. Future work may address
the integration of a grasping algorithm, the extension to include
motion of the robot arm during the handover, and the extension to
multiple robots for handovers tasks with large objects. Concerning
validation, future work may include integrating haptic feedback in the
interface, in order to provide tactile feedback of the handover task, as well
as performing test sessions on the ISS using the real NASA Astrobee
robots.

\section*{Acknowledgement}

This work was supported by the FCT Project LaRSYS (UIDB/50009/2020) and
the ANI P2020 project INFANTE (10/SI/2016).

\end{document}